\documentclass[sigchi,nonacm,xcolor=table,xcolor=xcdraw]{acmart}

\renewcommand\footnotetextcopyrightpermission[1]{} 
\RequirePackage{snapshot}

\setcopyright{rightsretained}

\acmConference[ACM SCHI]{ACM CHI Conference}{2019}{Glasgow, UK}
\acmYear{2019}
\copyrightyear{2019}
\acmPrice{15.00}

\usepackage{graphicx}
\usepackage{color}
\usepackage{booktabs}
\usepackage{colortbl}
\usepackage[sharp]{easylist}

\usepackage{subcaption}

\newcommand{\secref}[1]{\S\ref{sec:#1}}
\newcommand{\figref}[1]{Fig.~\ref{fig:#1}}
\newcommand{\tabref}[1]{Table~\ref{tab:#1}}
\newcommand{\vehiclename}{Human-Centered Autonomous Vehicle (HCAV)}

\newcommand{\website}{\url{https://hcai.mit.edu/hcav}}

\begin{document}

\title{Human-Centered Autonomous Vehicle Systems: Principles of Effective Shared Autonomy}

\author{Lex Fridman}
\affiliation{%
  \institution{Massachusetts Institute of Technology (MIT)}
}
\email{fridman@mit.edu}

\renewcommand{\shortauthors}{L. Fridamn et al.}
\renewcommand{\shorttitle}{Human-Centered Autonomous Vehicle Principles}

\begin{abstract}
Building effective, enjoyable, and safe autonomous vehicles is a lot harder than has historically been considered.  The reason is that, simply put, an autonomous vehicle must interact with human beings. This interaction is not a robotics problem nor a machine learning problem nor a psychology problem nor an economics problem nor a policy problem. It is all of these problems put into one. It challenges our assumptions about the limitations of human beings at their worst and the capabilities of artificial intelligence systems at their best. This work proposes a set of principles for designing and building autonomous vehicles in a human-centered way that does not run away from the complexity of human nature but instead embraces it. We describe our development of the \vehiclename{} as an illustrative case study of implementing these principles in practice.
\end{abstract}

%
%


\begin{teaserfigure}
  \includegraphics[width=\textwidth]{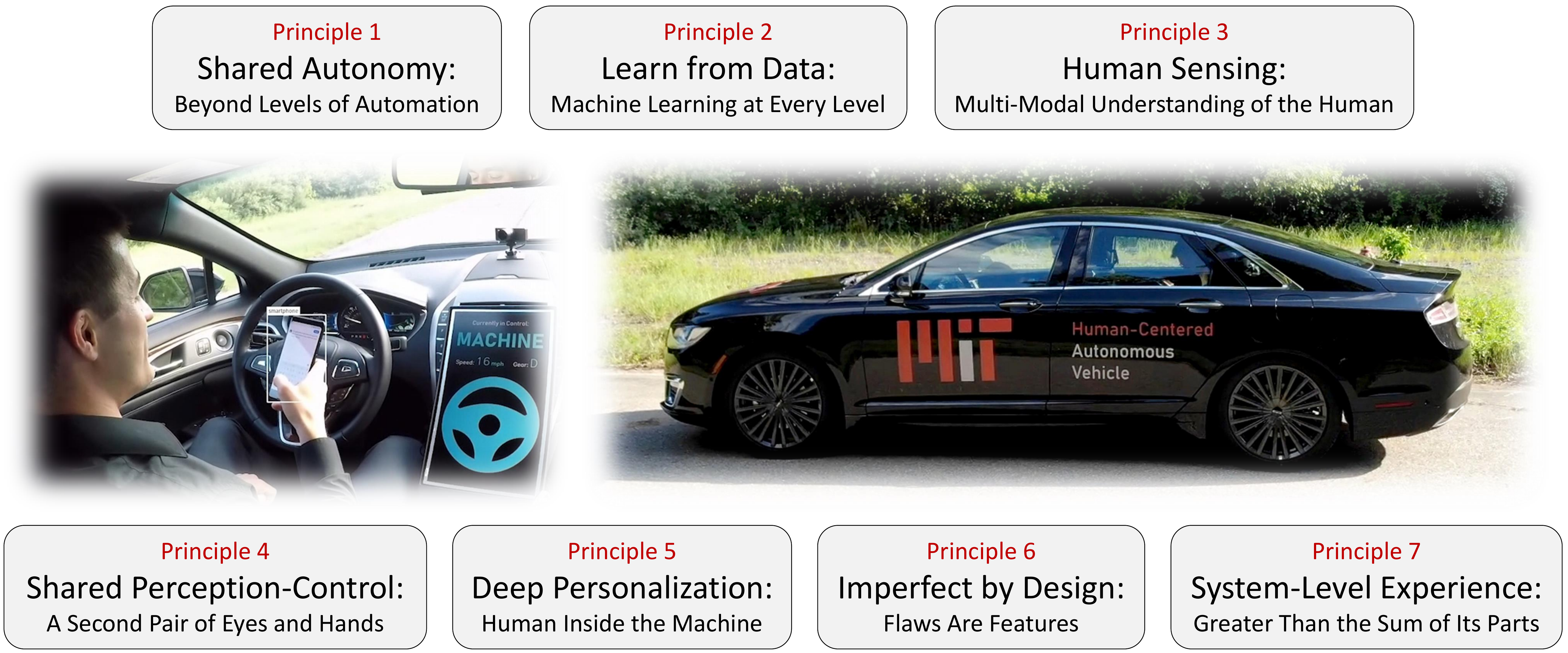}
  \caption{Principles of shared autonomy used for the design and development of the Human-Centered Autonomous Vehicle. }
  \label{fig:principles}
  \vspace{0.2in}
\end{teaserfigure}

\maketitle



\section*{Introduction}

Three ideas underlie the current popularly held view of autonomous vehicles:

\begin{enumerate}
\item The driving task is easy \cite{fridman2018avt,buehler2009darpa,mccarthy2018when}.
\item Humans are bad at driving \cite{davies2015look,vanderbilt2009traffic}.
\item Humans and automation don't mix well \cite{parasuraman1997humans,sheridan2002humans}.
\end{enumerate}

In contrast to this view, our work considers (1) that driving is in fact very difficult, (2) that humans are in fact
great drivers, and (3) that getting humans and artificial intelligence systems to collaborate effectively is an
achievable and worthy goal. In this light, we propose a human-centered paradigm for engineering shared autonomy systems
in the car that erase the boundary between human and machine in the way the driving task is experienced. Specifically,
we articulate seven principles of shared autonomy and discuss how we have applied these principles in practice during
the design, development, and testing of the \vehiclename. The statement and comparative details of each principle are
provided in the following sections. The goal of each principle is summarized here:

\begin{enumerate}
\item \textbf{Shared Autonomy:\\\textit{Beyond Levels of Automation}}\\
  \textit{Goal:} Motivate drawing a clear distinction shared autonomy and full autonomy.
\item \textbf{Learn from Data:\\\textit{Machine Learning at Every Level}}\\
  \textit{Goal:} Motivate the formulation of many software-based tasks as supervised machine learning problems thereby making them
  amenable to continuous improvement from data.
\item \textbf{Human Sensing:\\\textit{Multi-Modal Understand of the Human}}\\
  \textit{Goal:} Motivate the need for understanding the state of the driver both on a moment-by-moment basis and across
  hours, day, months, and years from multiple sensors streams.
\item \textbf{Shared Perception-Control:\\\textit{A Second Pair of Eyes and Hands}}\\
  \textit{Goal:} Motivate an approach for external perception, vehicle control, and navigation planning that seeks to
  inform and integrate the human driver into the driving experience even under highly-automated operation.
\item \textbf{Deep Personalization:\\\textit{Human Inside the Machine}}\\
  \textit{Goal:} Motivate adjusting the operation of the AI system to the individual driver to a degree of
  personalization where the resulting system more represents the behavior of the specific human driver than the generic
  background model of the vehicle as originally manufactured.
\item \textbf{Imperfect by Design:\\\textit{Flaws Are Features}}\\
  \textit{Goal:} Motivate redefining the goal for an autonomous vehicle as effective communication of flaws and
  limitations instead of flawless fully-autonomous operation.
\item \textbf{System-Level Experience:\\\textit{Greater Than the Sum of Its Parts}}\\
  \textit{Goal:} Motivate removing the focus on effectiveness of individual software components and instead focusing on
  the integrated shared autonomy experience.
\end{enumerate}

We implement the seven principles described in this work in a prototype vehicle shown in \figref{principles}. The
vehicle uses only cameras and primarily machine learning approaches to perform the driving scene perception, motion planning,
driver sensing, speech recognition, speech synthesis and managing the seamless two-way transfer of control via voice
commands and torque applied to steering wheel. Video demonstrations of the vehicle and the concepts described in this
paper are available online at \website.

\section{Shared Autonomy}\label{sec:shared-autonomy}

\newcommand{\principle}[1]{
  \noindent\fbox{%
    \parbox{0.96\columnwidth}{%
      \textbf{Principle:} #1
    }    
  }
}

\principle{Keep the human driver in the loop. The human-machine team must jointly maintain sufficient situation
  awareness to maintain control of the vehicle. Solve the human-robot interaction problem perfectly and the
  perception-control problem imperfectly.}

\vspace{0.1in}

The introduction of ever-increasing automation in vehicles over the previous decade has forced policymakers and safety
researchers to preemptively taxonomize automation in hopes of providing structure to laws, standards, engineering
designs, and the exchange of ideas. The six levels of automation (L0 to L5) is the result \cite{smith2013sae}. These
definitions have more ambiguously stated gray areas than clear, illuminating distinctions, and thus serve as no more
than reasonable openers for public discussion rather than a set of guidelines for the design and engineering of
automotive systems. We propose that shared autonomy and full autonomy are the only set of levels that provide
instructive guidelines, constraints, and goals for success. Moreover, they each provide a distinct set of challenge in
both kind and degree. These are illustrated in \tabref{levels}.

\begin{table}[]
\resizebox{\columnwidth}{!}{%
\begin{tabular}{l|c|c|}
\cline{2-3}
                                                 & \multicolumn{2}{c|}{\begin{tabular}[c]{@{}c@{}}Performance Level\\ Required\end{tabular}}                           \\ \cline{2-3} 
                                                 & \begin{tabular}[c]{@{}c@{}}Shared\\ Autonomy\end{tabular} & \begin{tabular}[c]{@{}c@{}}Full\\ Autonomy\end{tabular} \\ \hline
\multicolumn{1}{|l|}{Sensor Robustness \cite{baruah2018ranking}}          & \cellcolor[HTML]{FFFC9E}Good                              & \cellcolor[HTML]{9AFF99}Exceptional                     \\ \hline
\multicolumn{1}{|l|}{Mapping \cite{seif2016autonomous}}                    & \cellcolor[HTML]{FFFC9E}Good                              & \cellcolor[HTML]{9AFF99}Exceptional                     \\ \hline
\multicolumn{1}{|l|}{Localization \cite{levinson2010robust}}               & \cellcolor[HTML]{FFFC9E}Good                              & \cellcolor[HTML]{9AFF99}Exceptional                     \\ \hline
\multicolumn{1}{|l|}{Scene Perception \cite{cordts2016cityscapes}}           & \cellcolor[HTML]{FFFC9E}Good                              & \cellcolor[HTML]{9AFF99}Exceptional                     \\ \hline
\multicolumn{1}{|l|}{Motion Control \cite{bojarski2016end}}             & \cellcolor[HTML]{FFFC9E}Good                              & \cellcolor[HTML]{9AFF99}Exceptional                     \\ \hline
\multicolumn{1}{|l|}{Behavioral Planning \cite{paden2016survey}}        & \cellcolor[HTML]{FFFC9E}Good                              & \cellcolor[HTML]{9AFF99}Exceptional                     \\ \hline
\multicolumn{1}{|l|}{Safe Harbor}                & \cellcolor[HTML]{FFFC9E}Good                              & \cellcolor[HTML]{9AFF99}Exceptional                     \\ \hline
\multicolumn{1}{|l|}{External HMI \cite{fridman2017walk}}               & \cellcolor[HTML]{FFFC9E}Good                              & \cellcolor[HTML]{9AFF99}Exceptional                     \\ \hline
\multicolumn{1}{|l|}{Teleoperation* \cite{fong2001collaborative}}             & \cellcolor[HTML]{FFFC9E}Good                              & \cellcolor[HTML]{9AFF99}Exceptional                     \\ \hline
\multicolumn{1}{|l|}{Vehicle-to-Vehicle* \cite{harding2014vehicle}}        & \cellcolor[HTML]{FFFC9E}Good                              & \cellcolor[HTML]{9AFF99}Exceptional                     \\ \hline
\multicolumn{1}{|l|}{Vehicle-to-Infrastructure* \cite{milanes2012intelligent}} & \cellcolor[HTML]{FFFC9E}Good                              & \cellcolor[HTML]{9AFF99}Exceptional                     \\ \hline
\multicolumn{1}{|l|}{Driver Sensing \cite{fridman2016driver}}             & \cellcolor[HTML]{9AFF99}Exceptional                       & \cellcolor[HTML]{FFFC9E}Good                            \\ \hline
\multicolumn{1}{|l|}{Driver Communication}       & \cellcolor[HTML]{9AFF99}Exceptional                       & \cellcolor[HTML]{FFFC9E}Good                            \\ \hline
\multicolumn{1}{|l|}{Driver Collaboration}       & \cellcolor[HTML]{9AFF99}Exceptional                       & \cellcolor[HTML]{FFFC9E}Good                            \\ \hline
\multicolumn{1}{|l|}{Personalization}            & \cellcolor[HTML]{9AFF99}Exceptional                       & \cellcolor[HTML]{FFFC9E}Good                            \\ \hline
\end{tabular}%
}
\caption{Technology involved in (1) shared autonomy and (2) full autonomy approaches, including the required performance
  level of each techology for widespread deployement. General terms of ``good'' and ``exceptional'' are used to
  highlight the distinction of not solving the 1\% edge cases in the former case and having to solve them in the latter
  case. \textit{*Note:} Teleoperation, V2V, and V2I are not required technologies but if utilized would need to achieve
  the specified performance level. }
\label{tab:levels}
\end{table}

\subsection{Traditional Approach:}

The traditional approach to highly automated vehicles is to skip consideration of the human all-together and focus on
perfecting the mapping, perception, planning and other problems characterized by the ``exceptional'' performance
requirement under the \text{full autonomy} column of \tabref{levels}. Practically, considering current state-of-the-art
hardware and algorithmic capabilities, this approach puts a lot of emphasis on accurate high-definition mapping, robust
sensor suites, and conservative driving policies.

\subsection{Human-Centered Autonomous Vehicle Approach:}

As \tabref{levels} shows, the focus for HCAV is on the driver, from driver sensing (see \secref{human-sensing}) to
shared perception-control (see \secref{perception-control}) to communication and personalization (see
\secref{personalization}). Responsibility for the control of the vehicle remains with the driver, but depending on the
driver state, driver style, and prior joint-experience of the human and machine, much of the steering, acceleration, and
deceleration of the vehicle may be taken care of by the AI system. Tesla Autopilot, a current Level 2 system, is used on
average over for over 30\% of miles driven \cite{fridman2018apmiles}. Successful implementation of shared autonomy may
see over 50\% miles driven under machine control. In our implementation of HCAV, the vehicle is always able to maintain
take control with varying degrees of confidence, and the driver is always made aware of both the level of confidence and
the estimated risk from the perception system.

\section{Learn from Data}\label{sec:learn}

\principle{Every vehicle technology (see \tabref{levels}) should be data-driven. Each should collect edge-case data and
  continually improve from that data. The overall learning process should seek a scale of data that enables progress
  away from modular supervised learning formulations toward end-to-end semi-supervised and unsupervised learning
  formulations.}

\subsection{Traditional Approach:}

Traditional approach to vehicle autonomy at any level rarely involves significant machine learning except in a
specialized offline context of lane detection in Intel's Mobileye vision-based systems or infrared-camera based head
pose estimation in the GM's Super Cruise system. Tesla Autopilot has taken a further step in the software built on top
of the second version of its hardware toward converting more and more of the perception problem into a supervised
machine learning problem. Nevertheless, much of the control of the vehicle and the estimation of driver state (in the
rare cases it is considered) is engineered without utilizing large-scale data-driven methods and almost never updated in
an online learning process. In the case of fully autonomous vehicle undergoing testing today, machine learning is
primarily used for the scene understanding problem but not for any other aspect of the stack. Moreover, the amount of
data collected by these vehicles pales in scale and variability to that able to be collected by Level 2 vehicles.

\subsection{Human-Centered Autonomous Vehicle Approach:}

The data available in Level 2 vehicles for utilization within a machine learning framework is sufficiently expansive in
scale and scope and growing to capture varying, representative, and challenging edge cases. Shared autonomy requires
that both driver facing and driving scene facing sensory data is collected, mined, and used for supervised learning
annotation. In our implementation of HCAV, the driving scene perception, motion planning, driver sensing, speech
recognition, and speech synthesis are all neural network models that are regularly fine-tuned based on recently
collected driver experience data. In doing data collection, we do not focus on individual sensor streams but instead
consider the driving experience as a whole and collect all sensor streams together, synchronized via a real-time clock,
for multi-modal annotation. That is any annotation of the driving scene can be directly linked to any of the annotation
of the driver state. Performing annotation on synchronized sensor streams allows for easy transition from modular
supervised learning to end-to-end learning when the scale of data allows for it.

\newcommand{\cvfig}[2]{%
  \begin{subfigure}[t]{0.99\columnwidth}
    \includegraphics[width=\textwidth]{images/perception/#1.jpg}%
    \caption{#2}
    \label{fig:cv-#1}
  \end{subfigure}\hspace{0.06in}
}
\newcommand{\cvspace}{\vspace{0.08in}}

\begin{figure}[!ht]
  \centering
  \cvfig{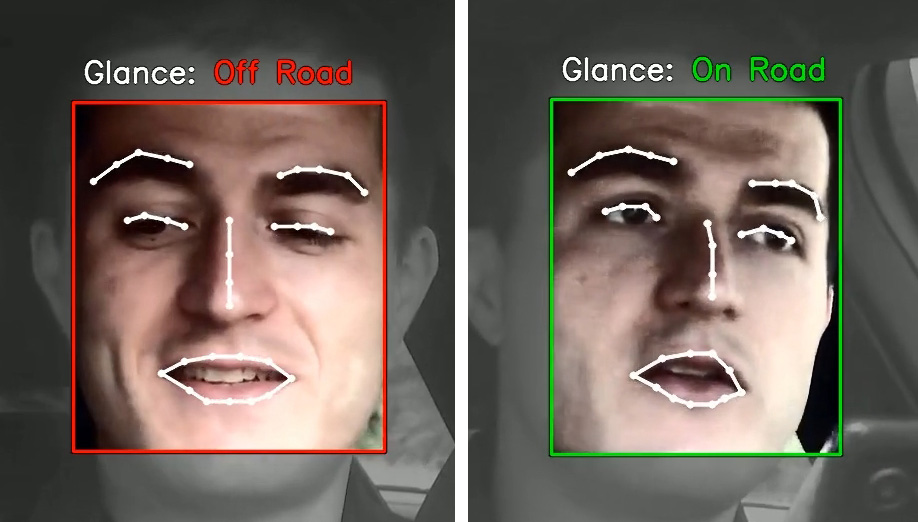}{Glance region classification.}
  \\\cvspace
  \cvfig{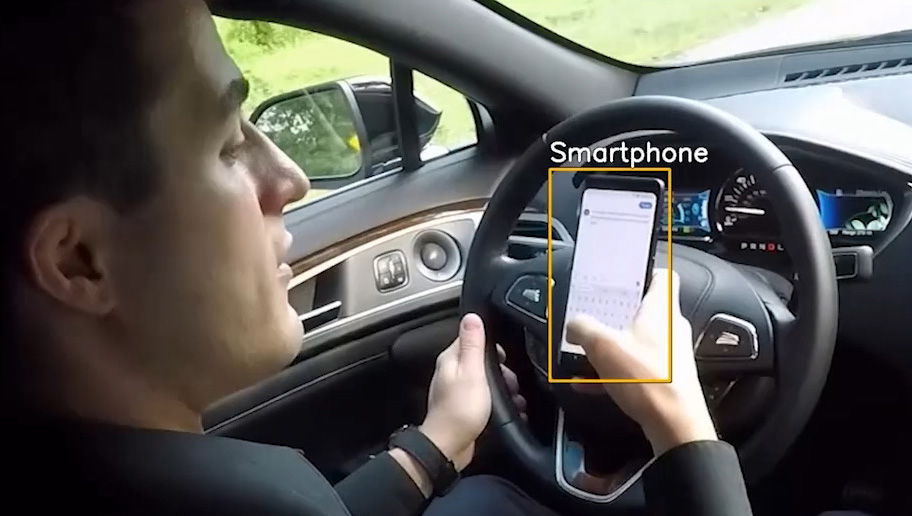}{In-cab object detection and activity recognition.}
  \\\cvspace
  \cvfig{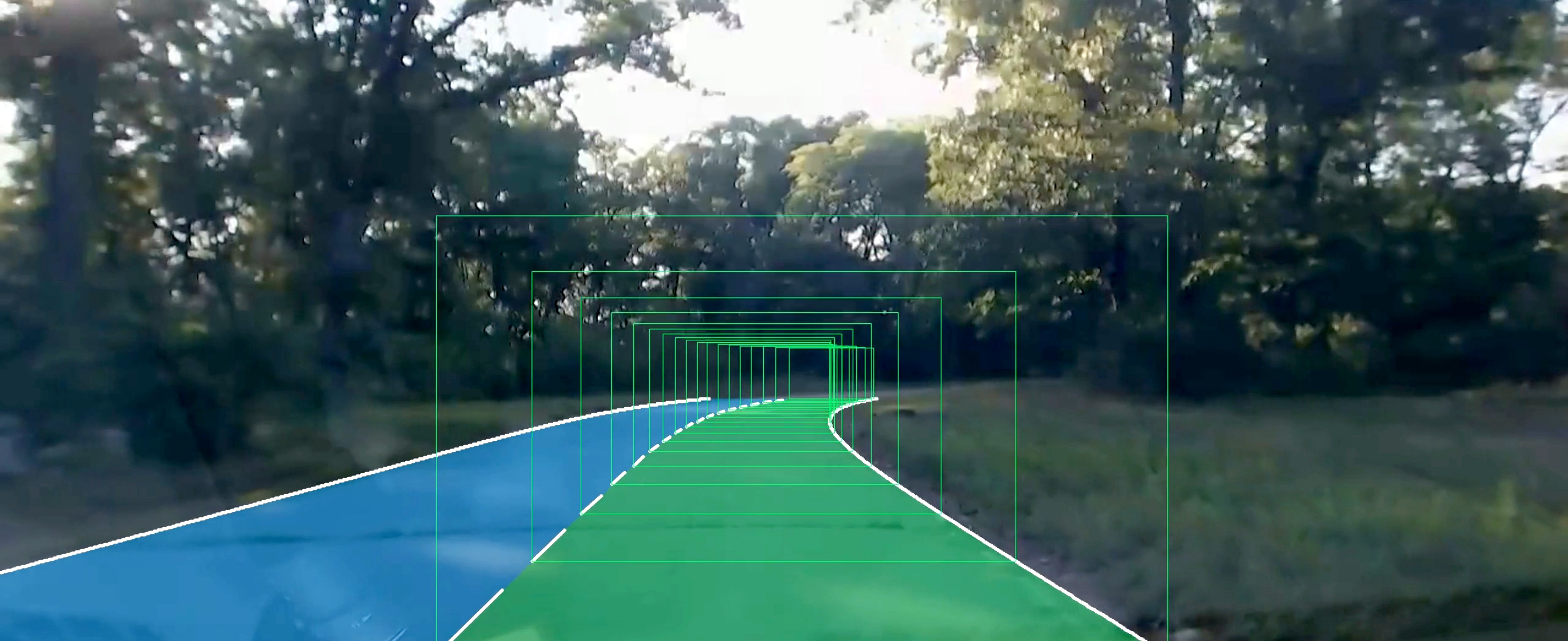}{Driveable area and lane detection.}
  \\\cvspace
  \cvfig{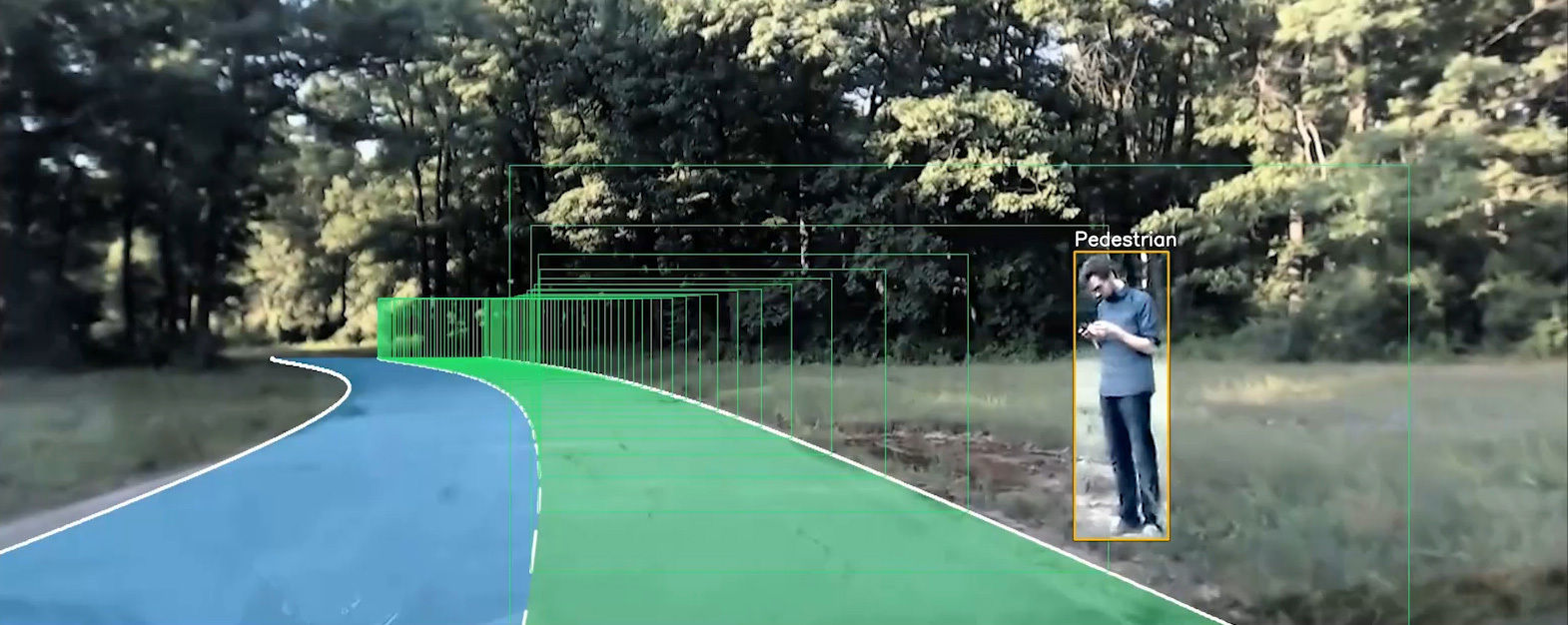}{Driving scene entity detection.}
  \caption{Perception tasks in our implemetation of HCAV. Visualization of the perception tasks integated to determine
    risk is shown in \figref{risk}.}
  \label{fig:cv}
\end{figure}

\section{Human Sensing}\label{sec:human-sensing}

\principle{Detect driver glance region, cognitive load, activity, hand and body position. Approach the driver state
  perception problem with equal or greater rigor and scale to the external perception problem.}

\vspace{0.1in}

Driver sensing refers to multi-modal estimation of overall physical and functional characteristics of the driver
including level of distraction, fatigue, attentional allocation and capacity, cognitive load, emotional state, and
activity. Typical driver gaze estimation\cite{murphy2009head} involves extraction of head and eye pose and estimation of
gaze or neural network based approaches that instead solve the gaze region classification problem
\cite{fridman2018avt}. Driver cognitive load estimation \cite{fridman2018cognitive} involves detection of working
memory load based on eye movement. Driver fatigue and drowsiness estimation \cite{wang2006driver} aims to detect arousal
from blink rates, eye movement, and body movement. In the driver state detection context, this is the most extensively
studied computer vision area. Driver emotion \cite{zeng2009survey,abdic2016driver} uses facial landmark configuration
and facial motion analysis. Physiological and audio sensors are often utilized in detection of affect.  Driver activity
recognition \cite{turaga2008machine,cheng2007multi} uses gaze patterns and movements of head, arms, hands, and
fingers. This includes detailed gesture recognition and broad activity type (i.e, smartphone texting) recognition.

\subsection{Traditional Approach:}

Driver sensing hardware and software capabilities are missing in almost all manual, semi-autonomous, and autonomous
vehicles being tested today. Exceptions include the GM Super Cruise system that has a camera on the steering wheel for
head tracking and Tesla Model 3 which has an in-cab camera that, to the best of our knowledge, is not currently utilized
for driver state estimation. Besides vision-based methods, crude low-resolution methods of driver sensing approaches include tracking steering
reversals as a proxy for driver drowsiness.

\subsection{Human-Centered Autonomous Vehicle Approach:}

Sensing the state of the driver is the first and most impactful step for building effective shared autonomy systems.
Automated methods for extracting actionable knowledge from monocular video of a driver have been actively studied for
over two decades in computer vision, signal processing, robotics, and human factors communities. The overarching goal
for these methods is to help keep the driver safe. More specifically, detection of driver state facilitates the more
effective study of how to improve (1) vehicle interfaces and (2) the design of future Advanced Driver Assistance Systems
(ADAS). With increasingly intelligent vehicle interfaces and the growing role of automation technology in the vehicle,
the task of accurate real-time detection of all aspects of driver behavior becomes critically important for a safe
personalized driving experience. Of special interest is the transition across different semi-autonomous driving modes
ranging from fully manual control to fully autonomous driving. The handoff in either direction of transition requires
that the vehicle has accurate information about the driver state. In our implementation of HCAV, we estimate driver
glance, cognitive load, and activity at 30 Hz.

\begin{figure*}[!ht]
  \centering
  \includegraphics[width=\textwidth]{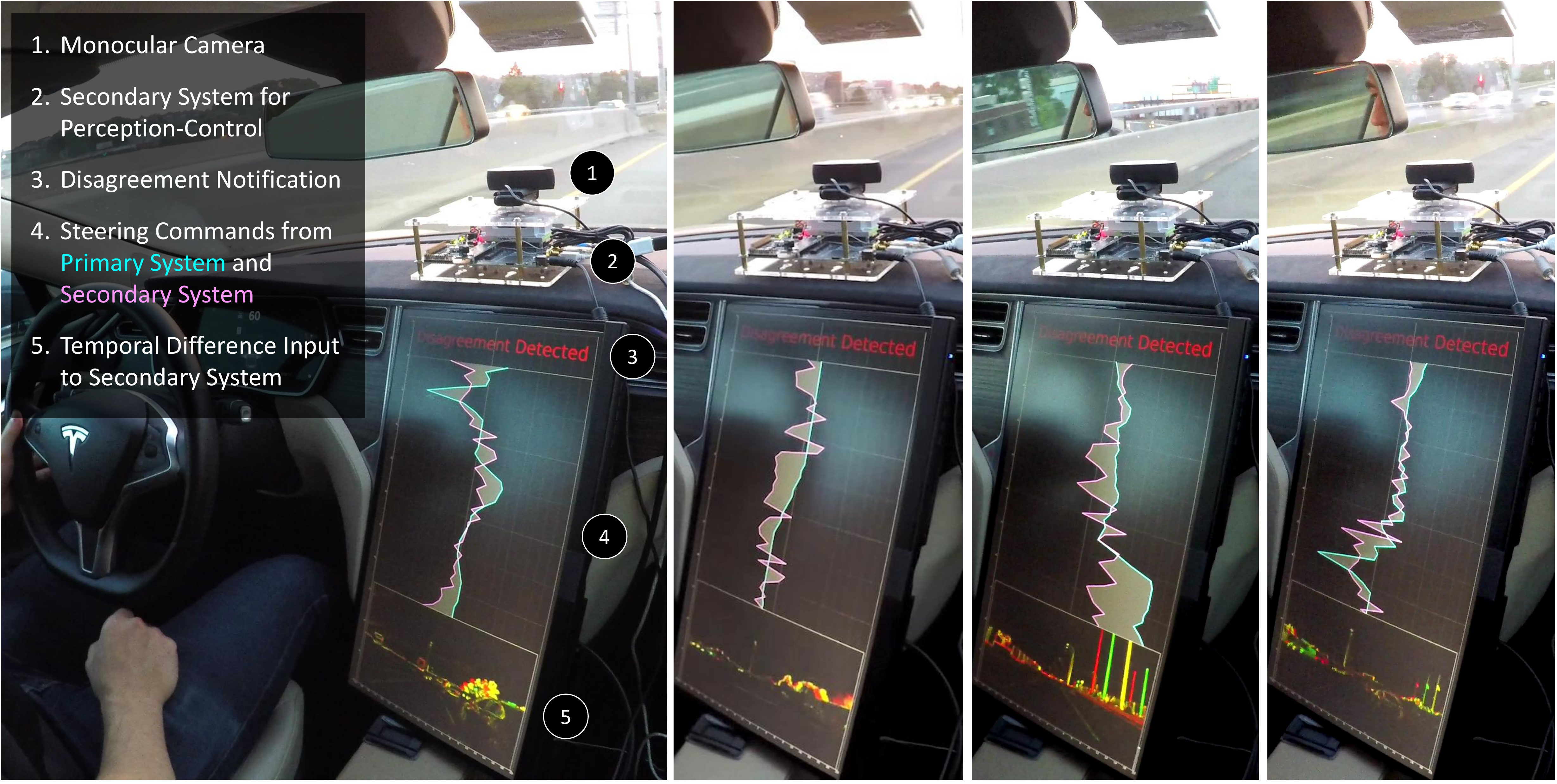}
  \caption{Implementation and evaluation of the arguing machines framework implemented in a Tesla Autopilot and our HCAV
    prototype vehicle. The technical details of the framework are detailed in \cite{fridman2018arguing}.}
  \label{fig:demo}
\end{figure*}

\section{Shared Perception-Control}\label{sec:perception-control}

\principle{Perform scene perception and understanding with the goal of informing the driver of the system capabilities
  and limitations, not with the goal of perfect black box safe navigation of the vehicle.}

\subsection{Traditional Approach:}

The goal of a fully autonomous vehicle is to perfectly solve the perception-control task, considering the human driver
an unreliable and unpredictable perturbation to the control problem. Removing the human being from the formulation of
the problem makes the problem appear better defined and thus seemingly more amenable to the type of approach that proved
successful in the DARPA Urban Challenge over 10 years ago \cite{buehler2009darpa}.

\newcommand{\riskfig}[2]{%
  \vspace{0.05in}
  \begin{subfigure}[t]{0.95\textwidth}
    \includegraphics[width=\textwidth]{images/perception/demo#1.jpg}%
    \caption{#2}
    \label{fig:cv-#1}
  \end{subfigure}\hspace{0.06in}
}
\newcommand{\riskspace}{\vspace{0.1in}}

\begin{figure*}[!ht]
  \centering
  \riskfig{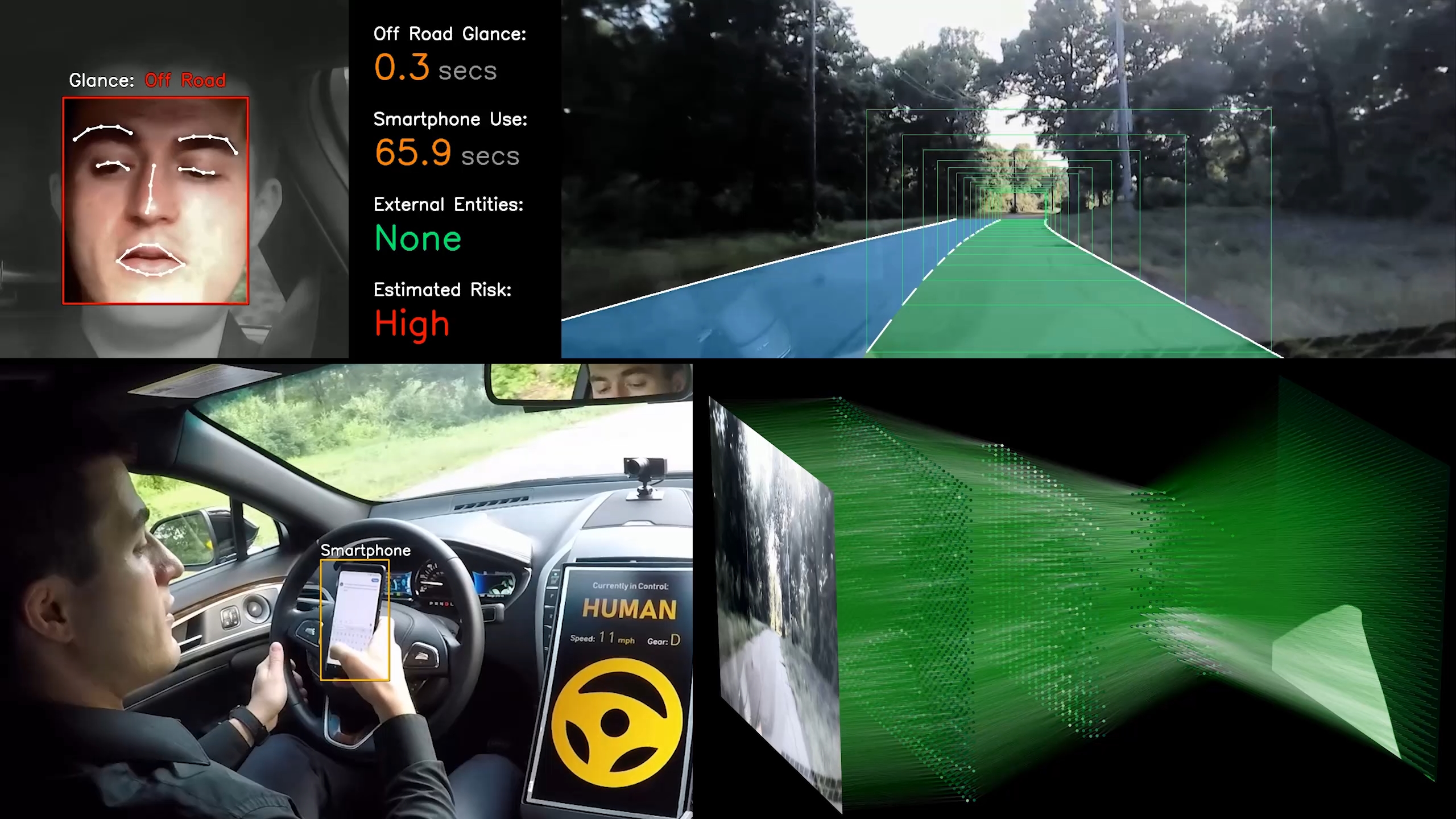}{Example of elevated risk under manual control during a period of frequent off-road glances to the smartphone.}\riskspace\\
  \riskfig{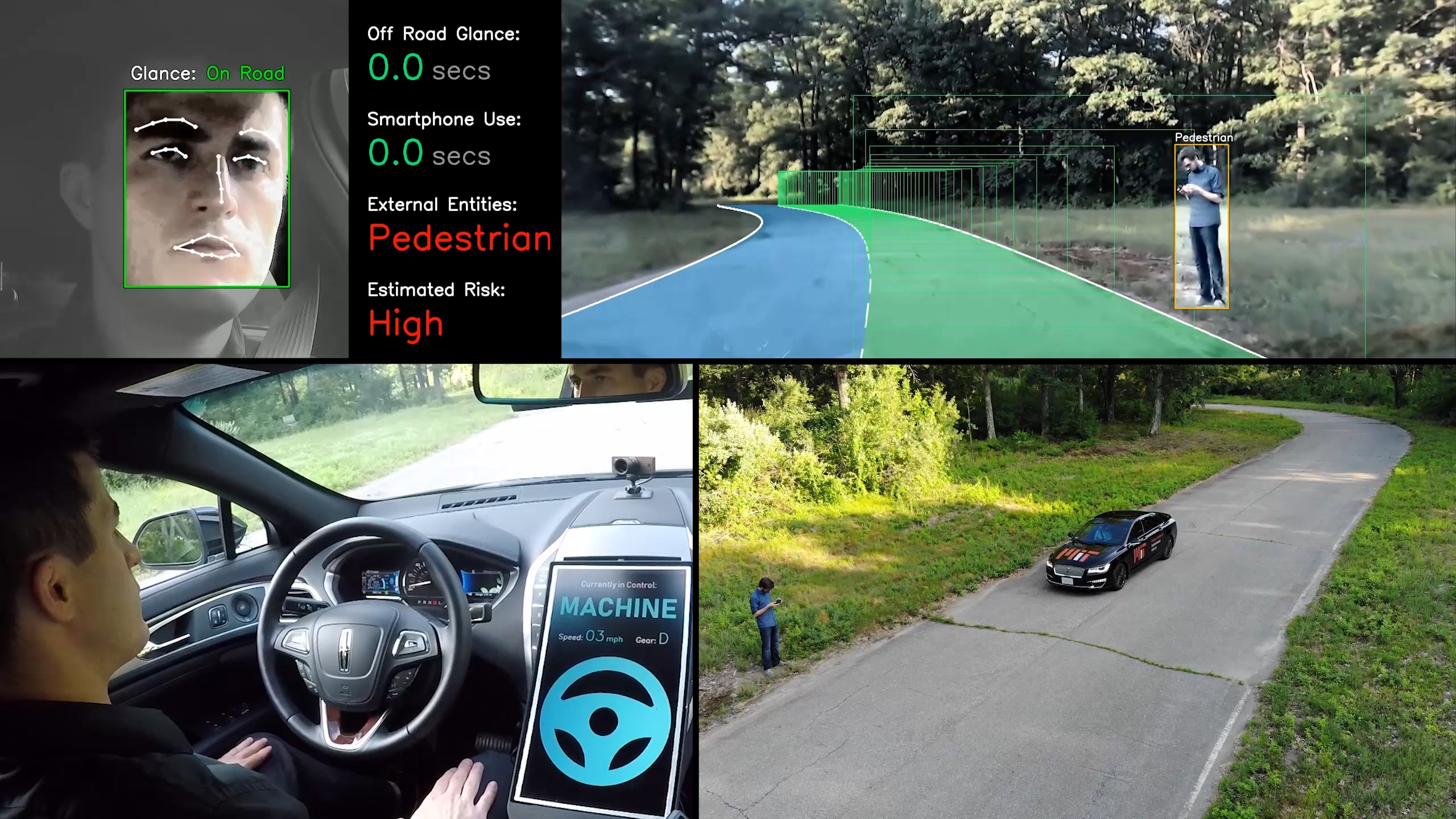}{Example of elevated risk under machine control in the presence of a pedestrian.}
  \caption{Examples of elevated risk computed by decision fusion of external and in-cab perception systems.}
  \label{fig:risk}
\end{figure*}

\subsection{Human-Centered Autonomous Vehicle Approach:}

Instead of decoupling the human driver from the loop of perception and movement planning, the human-centered approach by
definition seeks to integrate the human being. The goal of the perception task then becomes to support the human driver
with information about the external scene and more importantly about the \textit{limitations} of the perception
system. Communication of imperfection as discussed in \secref{imperfect} is the ultimate goal of the perception-control
task in the shared autonomy paradigm.

In our implementation of HCAV, there are several key algorithms that are designed around this principle. Examples are
shown in \figref{cv}. First, we visually communicate the degree of uncertainty in the neural network prediction,
segmentation, or estimation about the state of driving scene. Second, we integrate all the perception tasks in a
decision fusion step in order to estimate the overall risk in the scene as shown in \figref{risk}. The decision fusion
is across both internal and external facing sensors. Third, we are always doing imitation learning: using the steering
of the human driver when he or she is in control as training data for the end-to-end steering network. Fourth, we use
the end-to-end network as part of an \textit{arguing machines} framework (detailed in \cite{fridman2018arguing}) to
provide human supervision over the primary perception-control system as shown in our implementation of it in
\figref{demo}.

\section{Deep Personalization}\label{sec:personalization}

\principle{Every aspect of vehicle operation should be a reflection of the experiences the specific vehicle shares with
  the driver during their time together. From the first moment the car is driven, it is no longer like any other instance
  of it in the world.}





\subsection{Traditional Approach:}

The most common approach in designing and engineering automotive system is to do no personalization at all, except
minimally within the infotainment system interaction part of the driving experience. One of the ideas underlying such
engineering design is that people want a system to perform as expected, and in order to form correct expectations, the
behavior of the system should be consistent within and across instances of the vehicle. In the rare cases that the
system learns from the driver (i.e., current implementation of Tesla Autopilot), to the best of our knowledge the
learning is integrated into the overall knowledge base as part of fleet learning.

\subsection{Human-Centered Autonomous Vehicle Approach:}

One of the most important and novel principles underlying the HCAV concept is \textit{deep personalization}, or the
instant and continuous departure of system behavior from the background model to one that learns from the experience
shared by one specific instance of the system and one specific human driver. Part of the learning process is
\textit{fleet learning} where the data is used to update the background model (system behavior deployed to all
drivers). However, in terms of the overall system experience, the more impactful part is the \textit{individual
  learning} where the fine-tuned model controls the behavior of the system for only the one specific driver associated
with that model.

This approach has profound implications for several aspects of semi-autonomous driving. First, liability of many common
system ``failures'' rests on the human driver, much like a trainer of a horse is in part responsible for the behavior of
that horse when the horse is ridden. This concept is not a legal framework, but it is a way of creating an experience of
shared autonomy even when the vehicle is in control. It creates an operational and emotional closeness that we believe
is fundamental to successful human-machine collaboration in a safety-critical context of the driving task.

In our implementation of HCAV, this principle is applied in two areas: perception-control and communication. For motion
planning of the vehicle, we use imitation learning to adjust the objective function for choosing between the set of
generated candidate trajectories. For communication, we adjust the natural language dialogue system the vehicle uses to
inform the driver about changes in risk estimates and high-level shared autonomy decisions. The personalization is both in
the operation of the natural language generation and in the degree of personal feel. For example, the vehicle AI calls
the driver by their name and adjusts the tone of voice based on what was sufficient in the past to grab their attention.

\section{Principle 6: Imperfect by Design}\label{sec:imperfect}

\principle{Focus on communicating how the system sees the world, especially its limitations, instead of focusing on
  removing those limitations.}

\subsection{Traditional Approach:}

In the automotive context, for many reasons, engineering design is often focused on safety. Naturally, this leads to
goals formulated around minimizing frequency and magnitude of system failures. In other words, for autonomous driving,
perfection is the goal. The non-obvious side effect of such goals is that revealing imperfections and uncertainty often
becomes an undesirable design decision. The thought is: ``Why would you want to show imperfections when the system is
supposed to be perfect?'' There are, of course, legal and policy reasons for such design decisions as well that are in
large part outside the scope of this discussion.

\subsection{Human-Centered Autonomous Vehicle Approach:}


Rich, effective communication is the most essential element of designing and engineering artificial intelligence systems
in the shared autonomy paradigm. Within the context of communication, system imperfections are the most
information-dense content for exchanging and fusing models of the world between human and machine. Hiding system
uncertainty, limitations, and errors misses the opportunity to manage trust and form a deep bond of understanding with
the driver. In our view, it is one of the greatest design failings of prior attempts at implementing semi-autonomous systems.

In our implementation of HCAV, limitations of the system are communicated verbally and visually. We visualize the world
and the driver as the system seems them through the various algorithms mentioned in previous sections to help the driver
gain an intuition of system limits. As opposed to providing warnings or ambiguous signals about system uncertainty, we
found that simply showing the world as the car sees it is the most powerful method of communication. There are technical
challenges to this type of communication in that the process of visualization can often be more computationally
intensive than the perception-control and driver sensing algorithms themselves. However, we believe this is a
critically important problem to solve and thus deserves attention from the robotics, HRI, and HCI research communities.

\section{Principle 7: System-Level Experience}\label{sec:experience}

\principle{Optimize both for safety and enjoyment at the system level.}

\subsection{Traditional Approach:}

As described in \secref{imperfect}, one of the primary goal of the engineering design process in the automotive industry
is safety. Another major goal is lowering cost. This second goal tends to lead to modular, component-based design
thinking. The same pattern holds, for different reasons, in the design of artificial intelligence systems in robotics,
computer vision, and machine learning communities. Considering individual components (i.e., object detection) without
considering the overall experience (i.e., risk-based bi-directional transfer of control) allows to rigorously test the
individual algorithms and push the state-of-the-art of these algorithms forward. However, this process narrows the focus on
individual algorithms and not the experience of the overall system.

\subsection{Human-Centered Autonomous Vehicle Approach:}

The value of \textit{systems engineering} and \textit{systems thinking} has been extensively documented in literature
over the past several decades \cite{blanchard1990systems}. Nevertheless, this kind of thinking is rarely applied in the
design, testing, and evaluation of autonomous vehicles whether in their semi-autonomous or fully-autonomous
manifestations. As articulated in the other principles, both humans and AI systems have flaws, and only when the share
autonomy paradigm is considered at the system level do those flaws have a chance to be leveraged to become strengths.

\section{Conclusion}

It is difficult to predict which path to vehicle autonomy will prove successful both in the near-term and in the
long-term. Furthermore, it is not clear what success looks like. Our hope is that the goals of increased safety, an
enjoyable driving experience, and improved mobility can all be achieved without having to strike a difficult balance
between them. Moreover, we believe that while the shared autonomy approach is counter to the approach taken by most
people in automotive industry and robotics research community in the past decade, it nevertheless deserves serious
consideration. In the end, the choice rests on the question of whether solving the driving task perfectly is easier or
harder than perfectly managing the trust and attention of a human being. We believe this is far from a close case, and
this paper and our HCAV prototype vehicle is a serious attempt to consider shared autonomy as a path forward for
human-centered autonomous vehicle system development.

\section*{Acknowledgment}\label{sec:acknowledgement}

The authors would like to thank the team of engineers and researchers at MIT, Veoneer, and the broader driving and
artificial intelligence research community for their valuable feedback and discussions throughout the development of
this work. Support for this research was provided by Veoneer. The views and conclusions of authors expressed herein do not
necessarily reflect those of Veoneer.


\bibliographystyle{ACM-Reference-Format}
\bibliography{hcav}

\end{document}